\documentclass[11pt]{article}

\title{Phoneme recognition in TIMIT with BLSTM-CTC}

\author{
  Santiago Fern\'{a}ndez\footnotemark[1]
  \and
  Alex Graves\footnotemark[2]
  \and
  J\"{u}rgen Schmidhuber\footnotemark[1]\ \footnotemark[2]
}

\begin{document}
\maketitle

\renewcommand{\thefootnote}{\fnsymbol{footnote}}
\footnotetext[1]{IDSIA, Galleria 2, CH-6928 Manno-Lugano,
  Switzerland. E-mail: \texttt{\{santiago, juergen\}@idsia.ch.}}
\footnotetext[2]{TU Munich, Boltzmannstr. 3, D-85748 Garching, M\"{u}nchen,
  Germany. E-mail: \texttt{\{graves, juergen.schmidhuber\}@in.tum.de.}}
\renewcommand{\thefootnote}{\arabic{footnote}}

\begin{abstract}
  We compare the performance of a recurrent neural network with the best
  results published so far on phoneme recognition in the TIMIT
  database. These published results have been obtained with a combination
  of classifiers. However, in this paper we apply a single recurrent neural
  network to the same task. Our recurrent neural network attains an error
  rate of 24.6\%. This result is not significantly different from that
  obtained by the other best methods, but they rely on a combination of
  classifiers for achieving comparable performance.
\end{abstract}

\section{Introduction}
Spontaneous speech production is a continuous and dynamic process. This
continuity is reflected in the acoustics of speech sounds and, in
particular, in the transitions from one speech sound to another. As a
consequence, the boundaries between speech sounds are not clearly
defined. This fact significantly contributes to making segmentation and
labelling of speech data interrelated tasks. Because of this interrelation,
automatic speech recognition is best performed with methods such as hidden
Markov models (HMM) that do not require segmented data for development. On
the contrary, developing neural networks has traditionally relied on
segmented data. The objective functions require a network output target
value at every or specific time-steps in the data sequence. Connectionist
temporal classification (CTC) overcomes this limitation. CTC allows
developing neural network classifiers using a sequence of labels as the
desired output target~\cite{graves:icml2006}. Labels correspond to events
occurring in the input data sequence, such as phones in a speech data
stream. The number of labels in a target labelling is, therefore, typically
much shorter than the number of time-steps in the input data
sequence. Also, there is not timing information in a target labelling,
except for labels being in the same order in which events occur in the
input data sequence.

Recurrent neural networks are an interesting alternative to HMMs for speech
recognition. Their continuous internal state is naturally well suited for
modelling speech dynamics. Moreover, their capability to model data
dependencies has potential for modelling coarticulatory effects in
speech. In contrast, HMMs are built on a number of independence assumptions
about the data.

We showed in~\cite{graves:icml2006} that CTC-based recurrent neural
networks outperform state-of-the-art algorithms on phoneme recognition in
the TIMIT database. In contrast with the algorithms compared
in~\cite{graves:icml2006}, which rely on a single type of classifier to
perform the task, Glass' uses a committee-based
classifier~\cite{glass:csl2003}, whereas Deng \emph{et al}.'s combines the
scores from two related algorithms~\cite{deng:nips2005}. These two systems
achieved the best phoneme recognition rates published so far for TIMIT.  In
this paper, we compare the performance of a single CTC-based recurrent
neural network with that of Glass' and Deng \emph{et al}.'s systems. The
main differences with respect to the experimental setup used
in~\cite{graves:icml2006} are: first, the data are divided into training,
validation and test sets as described in~\cite{halberstadt:phdthesis1998},
and second, a standard set of 39 phonetic categories, instead of 61, is
used~\cite{lee:tassp1989}. This new experimental setup allows a direct
comparison of the three systems.

\section{Materials}
The DARPA TIMIT Acoustic-Phonetic Continuous Speech Corpus (TIMIT) contains
recordings of prompted English speech accompanied by manually segmented
phonetic transcripts~\cite{garafolo:ldc1993}. TIMIT contains a total of
6300 sentences, 10 sentences spoken by each of 630 speakers from 8 major
dialect regions of the United States.

For the experiments, the SA sentences were discarded and the remaining data
were split into a training set, a validation set and a test set according
to~\cite{halberstadt:phdthesis1998}. The training set contains 3696
sentences (462 speakers), the validation set contains 400 sentences (50
speakers) and the test set contains 192 sentences (24 speakers).

TIMIT transcriptions are based on 61 phones. Typically, 48 phones are
selected for modelling. Confusions among a number of these 48 phones are
not counted as errors. Therefore, results are presented for 39 phonetic
categories. We decided to train the network on transcriptions based on this
lexicon of 39 phones. The 61 categories were folded onto 39 categories as
described by Lee and Hon~\cite{lee:tassp1989}. This is shown in
table~\ref{tab:folding}.

\begin{table}
  \centering
  \begin{tabular}{l|l}
    aa  & aa, ao \\
    ah  & ah, ax, ax-h \\
    er  & er, axr \\
    hh  & hh, hv \\
    ih  & ih, ix \\
    l   & l, el \\
    m   & m, em \\
    n   & n, en, nx \\
    ng  & ng, eng \\
    sh  & sh, zh \\
    sil & pcl, tcl, kcl, bcl, dcl, gcl, h\#, pau, epi \\
    uw  & uw, ux \\
    --- & q \\
  \end{tabular}
  \caption{Folding the 61 categories in TIMIT onto 39 categories
    (from~\cite{lee:tassp1989}). The phones in the right column are
    folded onto their corresponding category in the left column (the
    phone 'q' is discarded). All other TIMIT phones are left intact.}
  \label{tab:folding}
\end{table}

Speech data was transformed into Mel frequency cepstral coefficients (MFCC)
with the HTK software package~\cite{young:cued2006}.  Spectral analysis was
carried out with a 40 channel Mel filter bank from 64\,Hz to 8\,kHz. A
pre-emphasis coefficient of 0.97 was used to correct spectral tilt. Twelve
MFCC plus the 0th order coefficient were computed on Hamming windows 25\,ms
long, every 10\,ms. Delta and Acceleration coefficients were added giving a
vector of 39 coefficients in total. For the network, the coefficients were
normalised to have mean zero and standard deviation one over the training
set.

The division of TIMIT into the three aforementioned data sets and the
presentation of results for 39 phones, was also adopted
in~\cite{glass:csl2003,deng:nips2005}. As acoustic features, Deng \emph{et
  al}. used frequency-warped LPC cepstra~\cite{deng:nips2005}, instead of
MFCC. For his part, Glass tried a number of variations and combinations of
MFCC, perceptual linear prediction (PLP) cepstral coefficients, energy and
duration~\cite{halberstadt:phdthesis1998,glass:csl2003}. Glass' system
built 61 models, one for each of the 61 phones in TIMIT, and results were
tabulated using the standard set of 39 phones.

\section{Method}

The method employed is the same described
in~\cite{graves:icml2006}. Briefly, phoneme recognition is performed with a
recurrent neural network. The long short-term memory recurrent neural
network (LSTM) was used because of its ability to bridge long time
delays~\cite{hochreiter:neuralcomp1997,gers:jmlr2002}. The hidden units in
an LSTM network are called memory blocks. Each memory block has one or more
memory cells controlled by an input, an output and a forget gate. When the
input gate is open incoming data is stored in the memory cell, and when the
output gate is open data stored in the memory cell is sent to the output
layer. The forget gate resets the memory cell. Gates can optionally have
access to the data stored in the memory cell (\emph{peephole}
connections). Gates and the memory block input are typically connected to
the same units in the network. These connections are trainable, thus the
behaviour of the gates is not pre-determined, but rather learned during
training.

For phoneme recognition, where both anticipatory and carry-over
coarticulatory effects are important, a bi-directional neural network is
suitable. The bi-directional LSTM
(BLSTM)~\cite{graves:nn2005,graves:icann2005} has two separate recurrent
hidden layers, both of them connected to the same input and output
layers. The \emph{forward} recurrent network is presented with sequential
data forward in time, from the beginning of the data sequence to time-step
$t$. The \emph{backward} recurrent network is presented with sequential
data backwards in time, from the end of the data sequence to time-step
$t$. At any time-step $t$, the network has access to all information in the
data sequence.

The BLSTM recurrent neural network was trained with the CTC algorithm using
the list of phones in the speech utterances as target
labellings~\cite{graves:icml2006}. Once the network has been trained, the
predicted labelling for a new speech utterance can be directly read from
its outputs. This method (best path decoding) is, however, not guaranteed
to find the most probable labelling. A second method (prefix search
decoding) consists in calculating the probabilities of successive
extensions of labelling prefixes, which can then be used to find the most
probable labelling. However, because this procedure is computationally
intensive, it was separately calculated for sections of the output
sequence. As a consequence, prefix search decoding is not guaranteed to
find the most probable labelling but, in practice, it generally outperforms
best path decoding~\cite{graves:icml2006}.

In the experiments reported in this paper, the BLSTM-CTC network had an
input layer of size 39, the forward and backward hidden layers had 128
blocks each, and the output layer was size 40 (39 phones plus blank). The
gates used a logistic sigmoid function in the range $[0, 1]$. The input
layer was fully connected to the hidden layer and the hidden layer was
fully connected to itself and the output layer. The total number of weights
was 183,080.

Training of the BLSTM-CTC network was done by gradient descent with weight
updates after every training example. In all cases, the learning rate was
$10^{-4}$, momentum was 0.9, weights were initialized randomly in the range
$[-0.1,0.1]$ and, during training, Gaussian noise with a standard deviation
of 0.6 was added to the inputs to improve generalisation. For prefix search
decoding, an activation threshold of 0.9999 was used
(see~\cite{graves:icml2006} for a description of this parameter).

Performance was measured as the normalised edit distance (label error rate;
LER) between the target label sequence and the output label sequence given
by the system.

Deng \emph{et al}.'s hidden trajectory models (HTM) are a type of
probabilistic generative model aimed at modelling speech dynamics and
adding long-contextual-span capabilities that are missing in hidden Markov
models (HMM)~\cite{deng:nips2005}. A thorough description of this system is
available in~\cite{yu:specom2006}. HTM uses a bi-directional filter to
estimate probabilistic speech data trajectories given a hypothesized phone
sequence. This estimate is then used to compute the model likelihood score
for the observed speech data. The search for the phone sequence with the
highest likelihood is performed with an A* based lattice search and
rescoring algorithm specifically developed for HTM.

Glass's system is a segment-based speech recogniser (as opposed to
frame-based recognisers) based on the detection of \emph{landmarks} in the
speech signal~\cite{glass:csl2003}. Acoustic features are computed over
hypothesized segments and at their boundaries. The standard decoding
framework is modified and extended to deal with this paradigm shift.

\section{Results}

Results are shown in table~\ref{tab:results}. Error rates include errors
due to substitutions, insertions and deletions with respect to the
reference transcription. Deng~\emph{et al}.'s best result was achieved with
a lattice-constrained A* search with weighted HTM, HMM, and language model
scores~\cite{deng:nips2005}. Glass's best results were achieved with many
heterogeneous information sources and classifier
combinations~\cite{glass:csl2003}. A single BLSTM-CTC recurrent neural
network attains an error rate of 24.6\%, which is not significantly
different from Deng~\emph{et al}.'s or Glass's best results. It is likely
that BLSTM-CTC can achieve improved performance when more sources of
information are added and when they are combined with other
classifiers. The results shown in table~\ref{tab:results} are the best
results reported in the literature on phoneme recognition in TIMIT.

\begin{table}
  \centering
  \begin{tabular}{ll}
    \hline \hline \\
    28.57\% & Deng~\emph{et al}.'s baseline HMM~\cite{deng:nips2005}\\ \hline\\
    25.17\%, s.e. 0.20\% & BLSTM-CTC (best path decoding) \\
    24.93\% & Deng~\emph{et al}.'s HTM-HMM \cite{deng:nips2005}\\ \hline\\
    24.93\% & Deng~\emph{et al}.'s HTM-HMM \cite{deng:nips2005}\\
    24.58\%, s.e. 0.20\% & BLSTM-CTC (prefix search decoding)\\
    24.4\% & Glass's committee-based classifier~\cite{glass:csl2003}\\
    \hline \hline
  \end{tabular}
  \caption{Error rates on TIMIT. Results for BLSTM-CTC are the average and
    standard error (s.e.) over 10 runs. On average, the networks were
    trained for 112.5 epochs (s.e. = 6.4). The horizontal lines divide the
    list of systems into groups performing significantly different than the
    networks. BLSTM-CTC with best path decoding is significantly different
    from Deng \emph{et al}.'s baseline HMM (two-sided t-test, $p<3\cdot
    10^{-8}$), from BLSTM-CTC with prefix search decoding ($p<0.05$) and
    from Glass's classifier ($p<0.004$). BLSTM-CTC with prefix search
    decoding is not significantly different from either Deng \emph{et
      al}.'s HTM-HMM or Glass's classifier.}
  \label{tab:results}
\end{table}

\section{Conclusions}
We have provided results for phoneme recognition with BLSTM-CTC using the
TIMIT database. The experiments use the same standard data sets and
phonetic inventory employed by the systems reportedly having the best
performance to date. Finally, we have compared BLSTM-CTC's performance to
that achieved by these
systems~\cite{glass:csl2003,deng:nips2005}. BLSTM-CTC achieves comparable
performance without relying on a combination of multiple classifiers. Also,
BLSTM-CTC makes fewer assumptions about the task domain.

\section*{Acknowledgments}
The authors would like to thank James R. Glass for providing the list of
speakers for each data set. This research was funded by SNF grant
200021-111968/1.

\end{document}